%% file: main.tex
\documentclass[runningheads]{llncs}

\usepackage{eccv}

\usepackage{eccvabbrv}

\usepackage{graphicx}
\usepackage{booktabs}

\usepackage[accsupp]{axessibility}  % Improves PDF readability for those with disabilities.

\usepackage{url}
\usepackage{balance}
\usepackage{array}
\usepackage{makecell}
\usepackage{xcolor}

\usepackage{dingbat}
\usepackage{multirow}
\usepackage{tabularx}
\usepackage{colortbl}
\usepackage{pifont}
\usepackage{hyperref}
\usepackage{mathtools}
\usepackage{subcaption}  % 提供 subfigure 环境（现代做法）
\usepackage{wrapfig}   % for wrapfigure environment
\usepackage[numbers,sort&compress]{natbib}
\DeclareUnicodeCharacter{2212}{$-$} % 把 U+2212 映到数学减号

\usepackage[utf8]{inputenc} % 新 TeXLive 通常默认 UTF-8，但写上更稳

\usepackage{hyperref}

\usepackage{orcidlink}

\begin{document}

% ---------------------------------------------------------------
% TODO REVIEW: Replace with your title
\title{Accelerating Diffusion Transformers with Gaussian Process Rectified Feature Cache} 

% TODO REVIEW: If the paper title is too long for the running head, you can set
% an abbreviated paper title here. If not, comment out.
\titlerunning{GP-Refiner}

% TODO FINAL: Replace with your author list. 
% Include the authors' OCRID for the camera-ready version, if at all possible.
% Remove the printed "and" before the last author in LNCS/ECCV style.
% Put before \begin{document}
\renewcommand{\lastandname}{}

\author{
Zhirong Shen\inst{1,3}\thanks{Equal contribution.} \and
Rui Huang\inst{1,3}$^\star$ \and
Chang Zou\inst{1,3} \and
Shikang Zheng\inst{1} \and
Jiacheng Liu\inst{1,4} \and
Peiliang Cai\inst{1} \and
Zhengyi Shi\inst{5} \and
Yaosong Du\inst{2} \and
Liang Feng\inst{6} \and
Xiaobing Tu\inst{2} \and
Jinkui Ren\inst{2} \and
Xiantao Zhang\inst{2}, \and 
Linfeng Zhang\inst{1}\thanks{Corresponding author.}
}

\authorrunning{Z. Shen et al.}

\institute{
Shanghai Jiao Tong University \and
Terminal Intelligent Computing Division, Alibaba Cloud \and
University of Electronic Science and Technology of China \and
Shandong University \and
Xiamen University \and
Fudan University\\[0.15em]
\email{\{2024080907011,huang\_rui\}@std.uestc.edu.cn}
\quad
\email{zhanglinfeng@sjtu.edu.cn}
}

\maketitle

\begin{abstract}
  Diffusion Transformers have become the dominant paradigm in generative AI, but their high computational costs severely hinder real-time applications. Prediction-based feature caching is widely used to accelerate diffusion transformers; however, as the number of steps increases, the deviation between its predictions and the reference full-compute trajectory gradually grows. An intuitive idea is to use an online regression model to dynamically correct this deviation, but it faces the issue of label data being unavailable during the acceleration process. This paper presents a statistical observation that the residuals between the features of full computation steps using caching methods and reference full-compute trajectory locally exhibit a zero-mean Gaussian distribution. By treating the features of full computation steps as noisy observations of reference features, the data acquisition problem is resolved. Based on this observation, a plug-and-play GP-Refiner correction framework is proposed. This method utilizes Gaussian Process Regression for correction and, leveraging the properties of GPR, introduces an uncertainty-adaptive computation strategy that triggers necessary full-computation calibration by monitoring the posterior variance in real time. Experiments demonstrate significant improvements across different models when combined with various state-of-the-art methods. Integrating the proposed framework with \textbf{TaylorSeer} reduces the computational load by \textbf{19.3\%} while improving PSNR by \textbf{0.9} dB and reducing LPIPS from \textbf{0.46} to \textbf{0.29}. Code is available in https://github.com/Aredstone/GP-Refiner.
\end{abstract}

\section{Introduction}

Diffusion models~\citep{DM,StableDiffusion,yang2025cogvideox} have become the dominant paradigm for visual generation, yet the adoption of Transformer-based architectures has drastically increased inference cost~\citep{chen2025s2guidancestochasticselfguidance,chen2025taming,blattmann2023SVD}. Existing \emph{training-free} feature caching techniques exploit temporal coherence along the diffusion trajectory by directly reusing intermediate features~\citep{ma2024deepcache,selvaraju2024fora}; however, under high acceleration ratios, the growing feature mismatch across long skip intervals leads to accumulated errors and noticeable quality degradation~\citep{liuTaylorSeer2025}. To overcome this bottleneck, \emph{forecasting-based} cache acceleration has recently emerged~\citep{zhengFoCa2025, fengHiCache2025}. Instead of pure reuse, it explicitly models the temporal evolution of intermediate representations to \emph{predict} future features from historical states, mitigating accuracy decay at large skip intervals and substantially reducing the number of expensive denoising-network forward evaluations while preserving generation fidelity.

Forecasting-based caching can substantially reduce the computation of diffusion sampling, but its prediction error typically accumulates with the skip interval, causing the generation trajectory to progressively drift away from the reference full-compute trajectory~\citep{Dicache, Liu2025SpeCa}. We aim to develop an adaptive rectification mechanism that requires no offline training and can be seamlessly plugged into existing forecasting-based caching accelerators, mapping the predicted features back to the reference full-compute trajectory under full-step sampling, denoted as $f_t$. However, during accelerated inference, once earlier steps have been skipped, even if we perform a full forward computation at the current timestep, we can only obtain $\hat{f}_t$, which is computed on an already drifted state and thus differs from the unbiased feature needed as supervision for training the rectifier.

\begin{figure}[t!]
    \centering
    \includegraphics[width=1\linewidth]{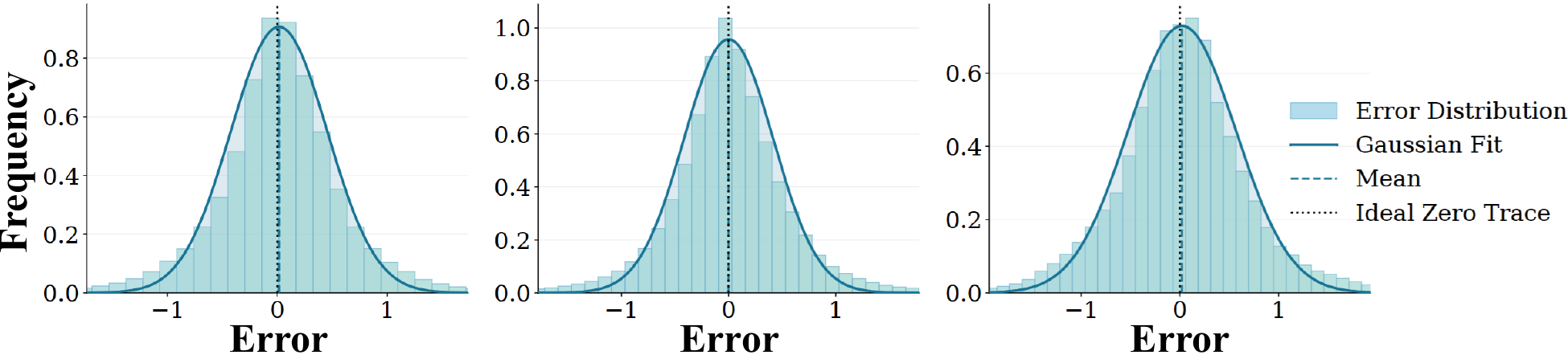}
    \caption{\textbf{Histogram of the error caused by the feature caching.} Across different images, the residuals are centered near zero and closely match a Gaussian fit, consistent with an approximately zero-mean Gaussian noise model.}
    \label{fig:obs1}
\end{figure}

\begin{figure}[t!]
    \centering
    \includegraphics[width=0.87\linewidth]{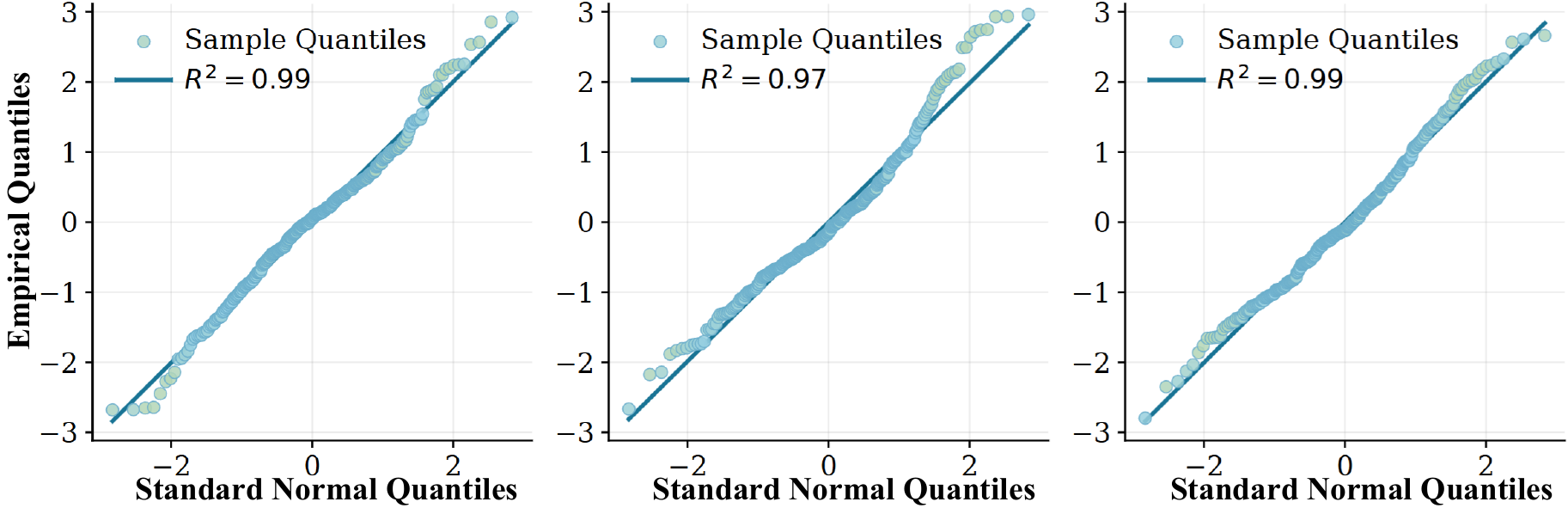}
    \hspace{1.2cm}
    \caption{\textbf{Q--Q plots of randomly projected feature residuals.}
The empirical quantiles closely follow a standard normal line ($R^2\!\approx\!0.97$--$0.99$), supporting an approximately zero-mean Gaussian residual model.}
    \label{fig:obs2}
\end{figure}

Our residual analyses of $f_t$ and $\hat{f}_t$ at both local and global levels show that, as illustrated in Fig.~\ref{fig:obs1} and Fig.~\ref{fig:obs2}, the residual between the full-compute-step features obtained under caching and the corresponding reference features exhibits a pronounced zero-mean Gaussian characteristic. Based on this observation, we can treat the full-compute-step feature $\hat{f}_t$ under caching as a Gaussian-noisy observation of the reference feature, thereby resolving the label acquisition challenge described above.

Motivated by the fact that Gaussian process regression (GPR) naturally accommodates observations corrupted bThe complete pipeline ofy zero-mean Gaussian noise, and that it matches the requirements of DiT cache acceleration, \emph{few-shot}, \emph{non-parametric}, and capable of fitting highly nonlinear mappings, we propose \textbf{GP-Refiner}, which leverages GPR to compensate for the bias in the baseline cache predictor. Moreover, we reformulate the problem as a joint task of one-step evolution and rectification: starting from a biased baseline prediction, we infer the unbiased reference value at the next moment. During inference, GP-Refiner takes the baseline predicted feature as input and performs GPR-based correction, driving the accelerated trajectory closer to the reference features while incurring almost no additional computation overhead.

Furthermore, we exploit another key property of GPR to enable an adaptive computation mechanism \citep{tang2024adadiff, chen2026jano}. By monitoring the posterior variance online, our algorithm can quantify the confidence of the current correction; when the uncertainty exceeds a predefined threshold, it forcibly triggers a full forward computation for state recalibration. This mechanism effectively establishes a dynamic balance between inference speed and generation quality.

\begin{figure}[t!]
    \centering
    \includegraphics[width=1\linewidth]{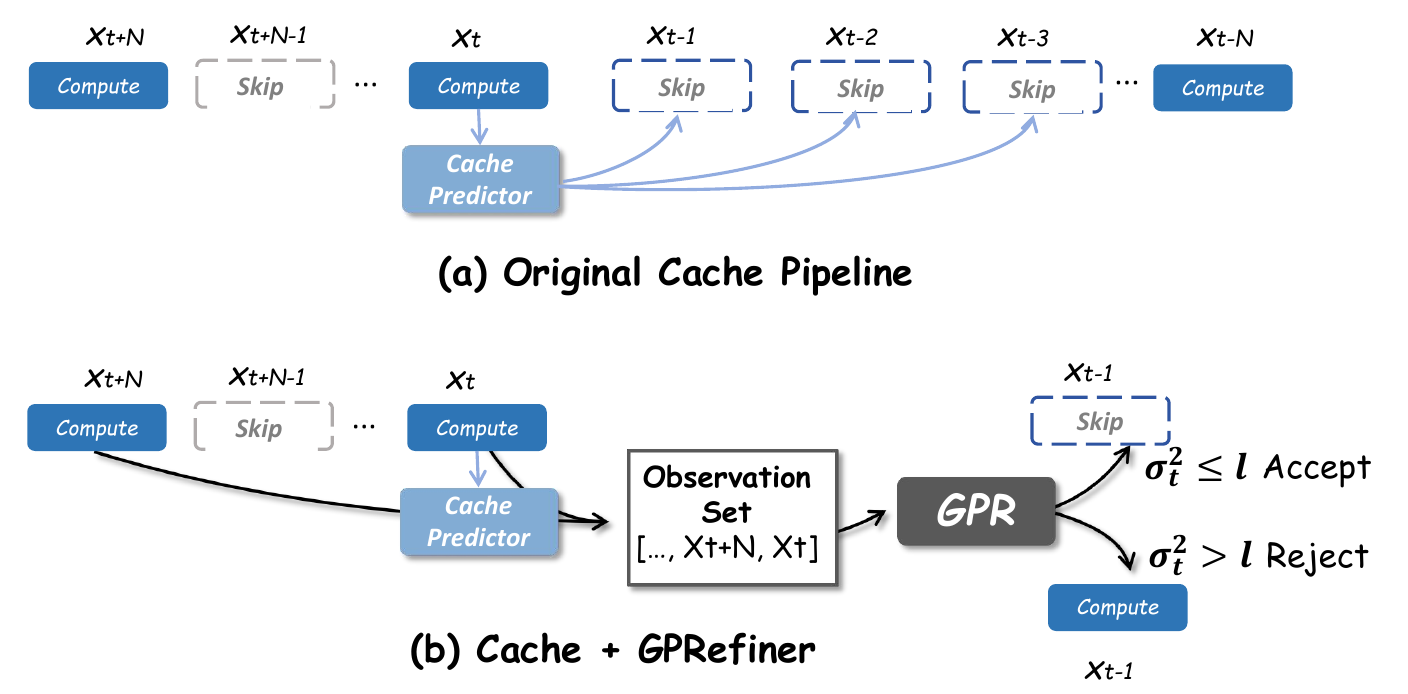}
    \caption{\textbf{Overall workflow of GP-Refiner.}
    \textbf{(a)} Original cache pipeline: features are computed at selected timesteps and cached, while skipped steps are filled by a cache predictor.
    \textbf{(b)} Cache + GP-Refiner: we collect an online observation set from full-compute steps, use GPR to rectify the predictor output, and trigger recomputation when the posterior variance exceeds a threshold ($\sigma_t^2 > l$).}
    \label{fig:main}
\end{figure}

The main contributions are divided into three aspects:
\begin{enumerate}
    \item \textbf{Noisy Observation Modeling from Statistical Evidence:}
    We reveal that the residuals between cached features and the corresponding reference (full-compute) features follow an approximately zero-mean Gaussian distribution. This allows us to treat full-compute-step features obtained under caching as Gaussian-noisy observations of the unbiased reference features, thereby resolving the supervision/label unavailability issue in accelerated inference and laying the foundation for regression-based bias correction.
    
    \item \textbf{GP-Refiner for Plug-and-Play Bias Rectification:}
    We propose \textbf{GP-Refiner}, which leverages Gaussian Process Regression to dynamically rectify the predictions of baseline cache forecasters. With nearly no additional compute overhead, GP-Refiner substantially improves generation fidelity under high acceleration.

    \item \textbf{Adaptive Re-computation via Posterior Uncertainty:}
    We introduce an adaptive computation strategy that quantifies prediction uncertainty via the Gaussian process posterior variance. When uncertainty exceeds a predefined threshold, we trigger a full forward evaluation to recalibrate the state, achieving a dynamic trade-off between inference speed and generation quality.
\end{enumerate}

\section{Related Work}

The computational burden of Diffusion Transformers (DiT)~\citep{huang2025diffusion,peebles2023dit,opensora} mainly arises from two factors: (i) the heavy Transformer computation when operating on high-resolution inputs~\citep{li_hunyuan-dit_2024, kong2024hunyuanvideo}, and (ii) the dozens of sequential denoising iterations required to preserve generation quality~\citep{wan_wan_2025,salimans2022progressive,zhao2024PAB}. Accordingly, most existing acceleration efforts target these two bottlenecks.

\subsection{Acceleration via Model Modification and Retraining}
A first line of research reduces inference cost by modifying the denoising model or compressing the sampling trajectory~\citep{structural_pruning_diffusion,yuan2024ditfastattn}, but requires additional retraining or fine-tuning~\citep{zhu2024dipgo,li2023FasterDiffusion}. Representative examples include structural pruning and token reduction to lower per-step Transformer compute~\citep{ , zhang2024tokenpruningcachingbetter, saghatchian2025cached}, as well as trajectory compression via consistency models and flow matching to enable few-step generation~\citep{song2023consistency, yan2024perflow, refitiedflow}. Despite their efficiency, these approaches are less plug-and-play for large, already-trained DiT models, which motivates our focus on training-free inference-time acceleration.

\subsection{Sampling Schedulers and Numerical Solvers}
A representative training-free acceleration direction is to improve the sampling scheduler by adopting more accurate numerical solvers, so that larger discrete step sizes can be used without sacrificing generation fidelity~\citep{ma2024l2c}. Typical examples include DDIM~\citep{songDDIM} and the DPM-Solver family~\citep{lu2022dpm, lu2022dpm++}, which leverage higher-order ODE-based solvers to stabilize coarse discretization. While effective in reducing the total number of steps, these methods still require a full forward pass of the denoising network at each step, leaving the per-step DiT computation largely unchanged and motivating complementary cache-based acceleration.

\subsection{Cache-Based Inference Acceleration}
\label{sec:rel_cache}
Cache-based acceleration is a training-free strategy that skips denoising-network computation at selected timesteps by exploiting the temporal coherence of intermediate representations along the diffusion trajectory. Existing methods can be broadly organized into three representative paradigms.
\subsubsection{\textbf{Cache-then-Reuse.}}
Reuse-based methods~\citep{ma2024deepcache,selvaraju2024fora} cache features at timestep $t$ and directly reuse them for subsequent skipped steps. They are simple and introduce minimal overhead, making them effective at modest speedups~\citep{zou2024DuCa,wimbauer2024cache}. However, feature similarity drops rapidly as the reuse interval increases, which causes pronounced drift and error accumulation under high acceleration ratios.
\subsubsection{\textbf{Cache-then-Forecast.}}
To support larger skip intervals, forecasting-based methods explicitly model the temporal evolution of features and predict future states from historical trajectories. TaylorSeer~\citep{liuTaylorSeer2025} pioneers this direction by treating feature evolution as a continuous-time process and extrapolating future features via Taylor-style high-order prediction. Subsequent works further improve long-horizon forecasting from different perspectives, including more stable polynomial or basis-function estimation and ODE-inspired formulations (e.g., HiCache~\citep{fengHiCache2025}, FoCa~\citep{zhengFoCa2025}). While substantially more robust than direct reuse, forecasting errors are still inevitable and can accumulate over long horizons, motivating the need for additional mechanisms to control drift.
\subsubsection{\textbf{Forecast-and-Correct.}}
% \noindent \textbf{(3) Forecast-and-Correct.}
To mitigate drift under aggressive skipping, recent work augments forecasting with feedback-driven correction mechanisms that introduce additional signals (e.g., verification or probing) to decide when to recompute and how to better track the denoising trajectory. SpeCa~\citep{Liu2025SpeCa} proposes a speculative caching framework that couples a lightweight predictor with low-cost verification, and rolls back or recomputes once the predicted trajectory exceeds a tolerance, improving robustness at high speedups. DiCache~\citep{Dicache} explores online alignment by leveraging probing signals to adaptively schedule caching and better match the underlying denoising dynamics. Overall, these designs go beyond pure forecasting by explicitly suppressing long-horizon error accumulation while retaining most of the computational gains from cache-based acceleration.

Overall, as the skip interval grows, both reuse-based and forecasting-based caching can accumulate errors and drift away from the true denoising trajectory. Existing corrective designs mitigate this drift using verification or probing signals to trigger recomputation, but they introduce extra overhead and remain limited for highly nonlinear feature errors. This motivates a \emph{training-free}, \emph{plug-and-play} module that performs \emph{nonlinear} correction while providing trustworthy \emph{uncertainty} or \emph{confidence} estimates for adaptive recomputation.

\section{Method}\label{sec:methodology}

\subsection{Preliminary}

\noindent \textbf{Diffusion Transformer (DiT). }
DiT is a diffusion-model backbone built entirely from Transformer layers and augmented with adaptive normalization.
Given an image at time $t$, the model operates on a sequence of patch tokens
$\mathbf{x}_t = \{x_i\}_{i=1}^{H \times W}$, where each $x_i$ encodes one image patch.
The network is a composition of $L$ Transformer blocks,
$\mathcal{G} = g_L \circ \cdots \circ g_2 \circ g_1$,
and each block combines self-attention, cross-attention, and a feed-forward subnetwork:
\begin{equation}
g_\ell(\cdot) = f_{\mathrm{MLP}}^{(\ell)} \left( f_{\mathrm{CA}}^{(\ell)} \left( f_{\mathrm{SA}}^{(\ell)}(\cdot) \right) \right),
\quad \ell = 1, \dots, L.
\end{equation}
Here, $f_{\mathrm{SA}}^{(\ell)}$ captures intra-token dependencies within the current latent sequence, $f_{\mathrm{CA}}^{(\ell)}$ conditions on external signals (e.g., text or timestep embeddings), and $f_{\mathrm{MLP}}^{(\ell)}$ provides non-linear mixing in token space.
Adaptive normalization layers modulate these components with timestep- and/or conditioning-dependent parameters, enabling the model to tailor its computations across the diffusion trajectory.

\noindent \textbf{Feature Caching for Diffusion Transformer.}
Temporal feature caching in DiT can be organized into two complementary paradigms: \emph{reuse-based} caching and \emph{forecasting-based} caching. 
In the reuse-based paradigm, a periodic schedule with interval $\mathcal{N}$ selects an anchor step $t$ at which all layer features are computed and stored, $\mathcal{C}(x_t^l) := \mathcal{F}(x_t^l)$ for $l \in \{0, \dots, L-1\}$. 
For the following $\mathcal{N}-1$ steps within the period, computation is skipped and features are reused,
\begin{equation}
\mathcal{F}(x_{t-k}^l) := \mathcal{C}(x_t^l), \quad k \in \{1, \dots, \mathcal{N} - 1\},
\end{equation}
yielding an approximate FLOPs reduction of $(\mathcal{N} - 1)/\mathcal{N}$ at the cost of an accumulated temporal mismatch as $\mathcal{N}$ grows. 
In the \emph{forecasting-based} paradigm, instead of directly reusing the anchor features, the skipped-step features are \emph{predicted} from cached temporal statistics that summarize local feature dynamics. Among forecasting-based methods, a representative approach is TaylorSeer~\citep{liuTaylorSeer2025}, which maintains the anchor feature together with its temporal finite differences up to order $m$,
\begin{equation}
\mathcal{C}(x_t^l) := \big\{\mathcal{F}(x_t^l), \Delta\mathcal{F}(x_t^l), \ldots, \Delta^m\mathcal{F}(x_t^l)\big\},
\end{equation}
and estimates feature at step $t-k$ via a truncated Taylor expansion around $t$,
\begin{equation}
\mathcal{F}_{\mathrm{pred},m}(x_{t-k}^l) = \mathcal{F}(x_t^l) + \sum_{i=1}^{m} \frac{\Delta^{i} \mathcal{F}(x_t^l)}{i! \mathcal{N}^{i}} (-k)^{i},
\end{equation}
where $\Delta^{i} \mathcal{F}(x_t^l)$ denotes the $i$-th temporal difference and the factor $\mathcal{N}^{-i}$ normalizes by the period length. This Taylor-series modeling explicitly captures short-horizon evolution of layer features and helps mitigate drift at larger skip intervals.

\noindent \textbf{Gaussian Process Regression.} 
In the context of probabilistic modeling, a Gaussian Process (GP) is defined as a collection of random variables, any finite number of which have a joint Gaussian distribution. Given a training set of input vectors $X = \{\mathbf{x}_i\}_{i=1}^n$ and a symmetric positive definite kernel function $k(\cdot, \cdot)$, we first define the training covariance matrix:
\begin{equation}
    \mathbf{K} = [k(\mathbf{x}_i, \mathbf{x}_j)] \in \mathbb{R}^{n \times n}.
\end{equation}
For a given test point $\mathbf{x}_*$, the covariance vector between the training points and the test point is defined as:
\begin{equation}
    \mathbf{k}_* = [k(\mathbf{x}_1, \mathbf{x}_*), \dots, k(\mathbf{x}_n, \mathbf{x}_*)]^{\top} \in \mathbb{R}^n.
\end{equation}
The prior self-covariance of the test point is expressed as the scalar:
\begin{equation}
    k_{**} = k(\mathbf{x}_*, \mathbf{x}_*).
\end{equation}
We assume that the observed outputs $\mathbf{y}$ are related to the latent function values through an additive i.i.d. Gaussian noise $\boldsymbol{\varepsilon} \sim \mathcal{N}(\mathbf{0}, \sigma_n^2 \mathbf{I})$. Consequently, the joint distribution of the observed training data $\mathbf{y}$ and the latent test value $g_*$ follows a multivariate Gaussian distribution:
\begin{equation}
    \begin{bmatrix}
    \mathbf{y} \\
    g_*
    \end{bmatrix}
    \sim \mathcal{N}\left( 
        \mathbf{0}, \, 
        \begin{bmatrix}
            \mathbf{K} + \sigma_n^2 \mathbf{I} & \mathbf{k}_* \\
            \mathbf{k}_*^{\top} & k_{**}
        \end{bmatrix} 
    \right).
\end{equation}
By conditioning the joint distribution on the observed data $\mathbf{y}$ and applying the properties of Schur complements, we derive the predictive distribution for the test point. The predictive mean, which represents the point estimate at $\mathbf{x}_*$, is given by:
\begin{equation}
    \mu_* = \mathbf{k}_*^{\top} \left( \mathbf{K} + \sigma_n^2 \mathbf{I} \right)^{-1} \mathbf{y}.
\end{equation}
Furthermore, the uncertainty associated with this prediction is quantified by the predictive variance:
\begin{equation}
    \label{gp_sigma_final}
    \sigma_*^2 = k_{**} - \mathbf{k}_*^{\top} \left( \mathbf{K} + \sigma_n^2 \mathbf{I} \right)^{-1} \mathbf{k}_*.
\end{equation}
These foundational expressions allow the model to provide both a regression value $\mu_*$ and a formal measure of confidence $\sigma_*^2$ for any unseen input.

\subsection{Observation}
In the accelerated generation process of diffusion models, an intuitive approach to correct the cumulative errors caused by approximation methods is to train a regression model for feature compensation. However, this faces a fundamental dilemma in practical applications: the problem of acquiring sample data. Due to the error accumulation from preceding timesteps, the features obtained during the acceleration process always deviate from the reference features on the standard denoising trajectory. This implies that we cannot collect accurate, real features online to serve as unbiased training samples.

To overcome this limitation, we conducted an in-depth analysis of the statistical properties of the feature deviation. Let $\hat{f}_t$ denote the feature obtained at a full computation step when using the caching method, and $f_t$ denote the true unbiased feature at the corresponding timestep. We define the residual between them as $\epsilon_t = f_t - \hat{f}_t$. Through extensive statistical testing, we observed a crucial phenomenon: the residual between the full computation step feature using the caching method and the reference feature exhibits significant characteristics of a zero-mean Gaussian distribution. We rigorously verified this phenomenon through experiments at both the micro and macro levels.

First, at the micro level, we extracted an arbitrary single residual vector $\epsilon_t$ and conducted a visual analysis of all its dimensional elements. As shown in Figure~\ref{fig:obs1}, the results indicate that the empirical probability density distribution of the elements within a single residual vector closely fits a standard Gaussian bell curve. This intuitively demonstrates that the elements of a single vector approximately follow a one-dimensional Gaussian distribution.

Meanwhile, we conducted multivariate Gaussian tests on multiple residual vectors at the macro level. We leveraged a core property from multivariate statistical theory: if a high-dimensional random vector follows a multivariate Gaussian distribution, its linear projection in any arbitrary direction must necessarily follow a one-dimensional Gaussian distribution. As shown in Figure~\ref{fig:obs2}, we linearly projected multiple residual vectors onto several randomly generated unit directions and plotted Q-Q plots for the projected one-dimensional data. Experimental calculations indicate that the scatter points of the projected data are highly concentrated along the theoretical diagonal of the Q-Q plot. Across $100$ random projections, the average correlation coefficient exceeded $0.98$, with the lowest correlation coefficient reaching $0.9424$. This rigorous quantitative test strongly confirms that the residual vector $\epsilon_t$ overall approximately follows a multivariate Gaussian distribution.

\subsection{Feature Correction Modeling via Gaussian Processes}

To rectify the cumulative errors during the accelerated generation of diffusion models, we reformulate the tasks of feature evolution and correction into a unified framework. The complete pipeline of GP-Refiner is illustrated in Figure~\ref{fig:main}, this section elaborates on how to infer unbiased ground-truth values for the next step by combining biased baseline predictions with historical observations.

\textbf{Formalization of Error Observation}
Based on the statistical analysis of feature deviations in Section 3.2, we observed a critical phenomenon: the residuals between the features generated by the baseline prediction method and the reference features exhibit a distinct Gaussian distribution. Based on this, we can mathematically reformulate the feature evolution process. Specifically, we treat the feature $\hat{f}_t$ obtained from a full-computation step as a ``noisy observation'' centered around the corresponding feature $f_t$ of the standard denoising process, perturbed by Gaussian noise. This relationship is formally expressed as:
\begin{equation}
\hat{f}_t = f_t + \epsilon
\end{equation}
where the residual $\epsilon$ is modeled as Gaussian noise with zero mean, denoted as:
\begin{equation}
\epsilon \sim \mathcal{N}(0, \sigma_n^2 \mathbf{I})
\end{equation}
This shift in modeling perspective is crucial. By integrating with Gaussian Process Regression (GPR), it elegantly bypasses the dilemma of being unable to obtain large-scale unbiased training samples in acceleration scenarios.

\textbf{Dynamic Construction of Baseline Methods and Observation Sets}
Guided by the aforementioned observation, the model dynamically maintains an observation set $\mathcal{D}_t$ during the inference stage to collect the data for GPR. The specific acquisition and construction mechanism of this set is as follows:

% \begin{figure}[t!]
%     \centering
%     \includegraphics[width=1\linewidth]{fig/main2.pdf}
%     \caption{\textbf{The structure of the feature correction and adaptive computation} module based on Gaussian Process Regression.}
%     \label{fig:main2}
% \end{figure}

When the model triggers a full network forward pass at a historical time step $k$, we first obtain the accurate full-computation feature $\hat{f}_{k}$. Subsequently, this value is used to update the state of the baseline prediction method. Utilizing the updated state and the extrapolative capability of the baseline method, we ``predict'' the feature of the previous step. We define the predicted feature output by the baseline method as:
\begin{equation}
\hat{p}_{k - 1} = \text{BaseMethod}(k - 1)
\end{equation}
Following this, we pair the baseline prediction for the previous step $\hat{p}_{k - 1}$ with the current full-computation feature $\hat{f}_{k}$ to form a data pair $(\hat{p}_{k - 1}, \hat{f}_{k})$, which is then formally added to the observation set. For consistency in subsequent mathematical expressions, we denote the $i$-th data pair added to the observation set as $(\hat{p}_{\tau_i - 1}, \hat{f}_{\tau_i})$, where $\hat{p}_{\tau_i - 1}$ represents the baseline predicted feature and $\hat{f}_{\tau_i}$ represents the noisy observation of the local ground-truth feature.

Before reaching the current time step $t$, if the system has collected $n$ historical observation pairs, the observation set can be formally defined as:
\begin{equation}
\mathcal{D}_t = \{(\hat{p}_{\tau_i - 1}, \hat{f}_{\tau_i})\}_{i=1}^n
\end{equation}
Implementing this pairing mechanism between adjacent steps is of significant theoretical importance. It strictly confines the focus of error correction to the local residuals of single-step evolution. This local pairing strategy effectively reduces the fitting pressure on the model, allowing it to concentrate on short-term evolutionary deviations rather than attempting to correct highly non-linear, long-span error accumulations.

\textbf{Introduction and Derivation of Gaussian Process Regression}
With the construction of the observation set clearly defined, we formally introduce the GPR framework. Distinct from traditional time-series fitting, we treat the baseline predicted features $\hat{p}$ as the input space and the full-computation features $\hat{f}$ as noisy observations within that space. This approach allows the model to learn the mapping residuals from baseline predictions to unbiased reference features, thereby calibrating the feature trajectory in real-time during inference.

To perform numerical inference, we first extract information from the observation set $\mathcal{D}_t$ to construct the core mathematical entities. First is the observation vector $\mathbf{Y}$, which aggregates the ground-truth feature values captured in all historical full-computation steps:
\begin{equation}
\mathbf{Y} = [\hat{f}_{\tau_1}, \hat{f}_{\tau_2}, \dots, \hat{f}_{\tau_n}]^T
\end{equation}
Correspondingly, we construct the prediction input matrix $\mathbf{X}$, which contains the associated baseline predicted features:
\begin{equation}
\mathbf{X} = [\hat{p}_{\tau_1 - 1}, \hat{p}_{\tau_2 - 1}, \dots, \hat{p}_{\tau_n - 1}]^T
\end{equation}
In this context, $\mathbf{X}$ serves as the feature input for the regression task, while $\mathbf{Y}$ represents the target observations we aim to approximate.

Next, we construct the training covariance matrix (Gram matrix) $\mathbf{K}$. Each element $K_{ij}$ is computed via a kernel function $k(\cdot, \cdot)$:
\begin{equation}
K_{ij} = k(\hat{p}_{\tau_i - 1}, \hat{p}_{\tau_j - 1})
\end{equation}
Since we defined the full-computation feature $\hat{f}$ as a noisy observation with variance $\sigma_n^2$, we utilize the modified covariance matrix $(\mathbf{K} + \sigma_n^2 \mathbf{I})$ for the matrix inversion during inference.

Simultaneously, to characterize the correlation between the current predicted feature $\hat{p}_t$ to be corrected and the historical observations, we construct the test covariance vector $\mathbf{k}_*$:
\begin{equation}
\mathbf{k}_* = [k(\hat{p}_t, \hat{p}_{\tau_1 - 1}), k(\hat{p}_t, \hat{p}_{\tau_2 - 1}), \dots, k(\hat{p}_t, \hat{p}_{\tau_n - 1})]^T
\end{equation}

Based on the conditional distribution theory of Gaussian Processes, we can derive the corrected unbiased feature prediction mean $\mu_t$ for the current step $t$:
\begin{equation}
\mu_t = \mathbf{k}_*^T (\mathbf{K} + \sigma_n^2 \mathbf{I})^{-1} \mathbf{Y}
\end{equation}
The framework simultaneously provides the prediction variance $\sigma_t^2$ as follows:
\begin{equation}
\sigma_t^2 = k(\hat{p}_t, \hat{p}_t) - \mathbf{k}_*^T (\mathbf{K} + \sigma_n^2 \mathbf{I})^{-1} \mathbf{k}_*
\end{equation}

% \begin{figure}[t!]
%     \centering
%     \includegraphics[width=1\linewidth]{fig/dy.pdf}
%     \caption{\textbf{Visualization of the feature L2 error.} The L2 distance between the approximated features and the reference full-compute features.}
%     \label{fig:dy}
% \end{figure}

\subsection{Adaptive Computation via Uncertainty}

The GP-Refiner framework not only rectifies feature deviations but also provides a real-time quantification of inference trajectory uncertainty through the predicted variance $\sigma_t^2$. To overcome the limitations of fixed computation intervals, we introduce a preset variance threshold $l$, which allows the model to dynamically select the computation mode based on whether the current uncertainty exceeds a tolerable range.

At each time step $t$, the system first derives the predicted variance $\sigma_t^2$ of the current corrected feature via the Gaussian Process. This variance intuitively reflects the confidence level of the current baseline prediction $\hat{p}_t$ relative to the known feature manifold (i.e., the observation set $\mathcal{D}_t$). By setting an activation threshold $l$, we define the feature acquisition logic as the following piecewise decision process:
\begin{equation}
f_t^* = 
\begin{cases} 
\text{FullCompute}(t), & \text{if } \sigma_t^2 > l \\
\mu_t, & \text{if } \sigma_t^2 \le l
\end{cases}
\end{equation}
where $f_t^*$ represents the feature used for the current inference step. The selection of the threshold will be detailed in the appendix. It exhibits strong robustness and does not significantly impact the results when kept within a certain range. If the predicted variance $\sigma_t^2$ exceeds the preset threshold, indicating that the untrustworthiness of the baseline prediction has reached its limit, the system immediately triggers a full computation to obtain the accurate feature; otherwise, it employs the corrected mean $\mu_t$ from the Gaussian Process for rapid inference.

This threshold-based activation mechanism ensures that full-computation steps are naturally clustered during stages of the denoising trajectory where non-linearity is strongest and error sensitivity is highest, effectively finding an optimal balance between inference speed and image fidelity.

\section{Experiments}\label{sec:experiments}

\input{tables/FLUX-Metric}

\input{tables/Qwen-Metric}

\subsection{Experiment Settings}

\begin{figure}
    \centering
    \includegraphics[width=1\linewidth]{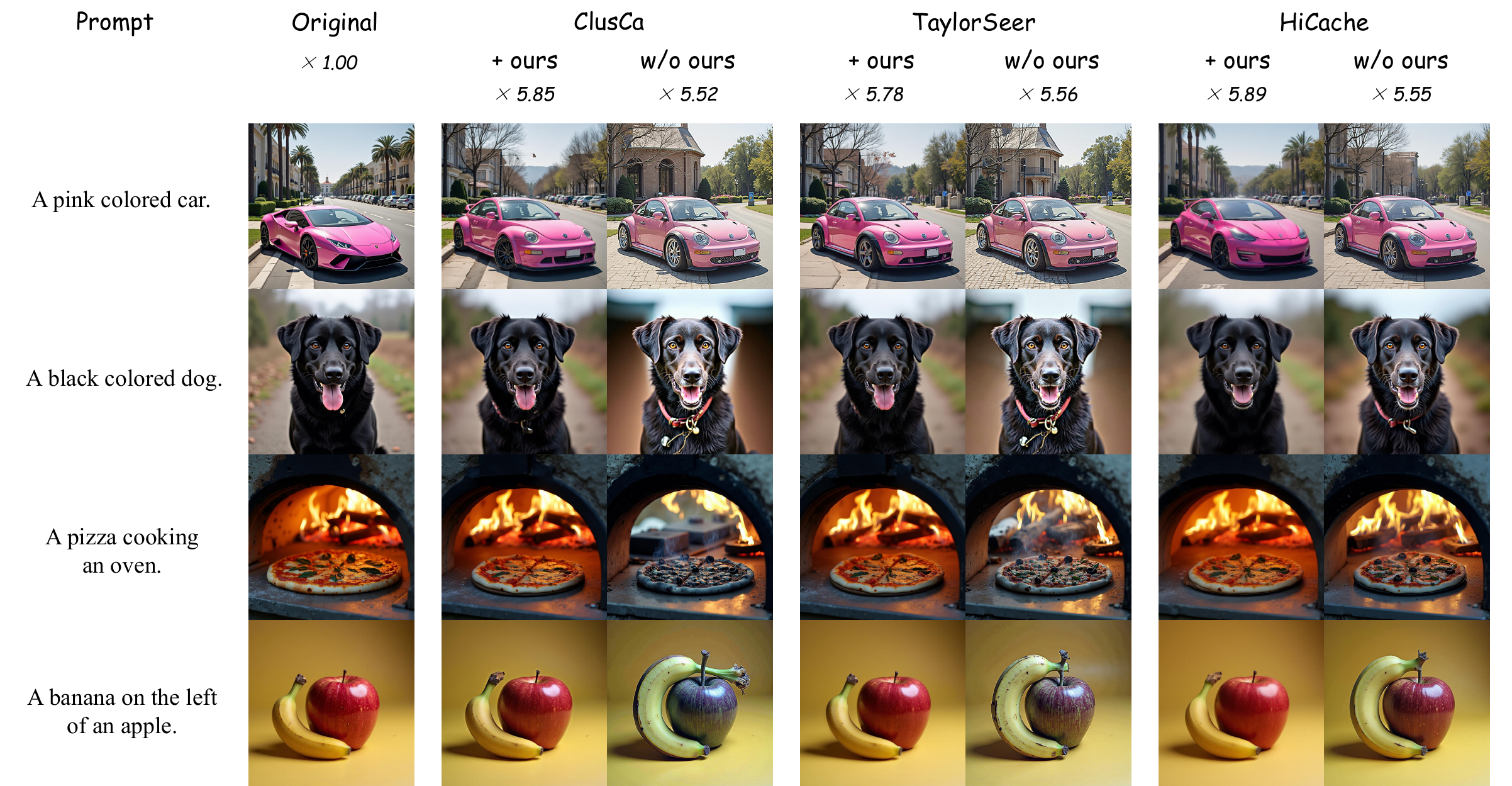}
    \caption{\textbf{Visual comparison} of generated images before and after combining different baseline cache methods with GP-Refiner.}
    \label{fig:compare}
\end{figure}

% \noindent \textbf{Model Configurations.}
% To validate the effectiveness of our method and its generality as a cache-rectification plugin, we conduct experiments on two commonly used text-to-image generators: \textbf{Qwen-Image}~\citep{Wu2025QwenImageTR} and \textbf{FLUX.1-dev}~\citep{flux2024}. As acceleration backbones, we integrate our rectifier with three representative caching methods: \textbf{TaylorSeer}~\citep{liuTaylorSeer2025}, \textbf{ClusCa}~\citep{zheng2025compute}, and \textbf{HiCache}~\citep{fengHiCache2025}. We instantiate the Gaussian-process rectifier with an \emph{angular RBF} kernel, and for local regression, we use a fixed FIFO window of h=10 most recent features.

% \noindent \textbf{Evaluation and Metrics. }
% For text-to-image generation, we employ 200 DrawBench~\citep{saharia2022drawbench} prompts as inference inputs. We evaluate the generated images using SSIM, PSNR, and LPIPS, and report the mean over all prompts.

\noindent \textbf{Model Configurations.}
To validate the effectiveness of our method and its generality as a cache-rectification plugin, we conduct experiments on commonly used text-to-image generators: \textbf{Qwen-Image}~\citep{Wu2025QwenImageTR} and \textbf{FLUX.1-dev}~\citep{flux2024}. Both base models operate at 50 inference steps. To further demonstrate robustness under extreme low-step regimes, we also include the Qwen-Image-Lightning variant running at 8 steps. As acceleration backbones, we integrate our rectifier with three representative caching methods: \textbf{TaylorSeer}~\citep{liuTaylorSeer2025}, \textbf{ClusCa}~\citep{zheng2025compute}, and \textbf{HiCache}~\citep{fengHiCache2025}. 

For comprehensive comparison, we benchmark against standard step reduction and recent caching baselines including \textbf{FORA}, \textbf{ToCa}, \textbf{DuCa}, and \textbf{TeaCache}. We instantiate the Gaussian-process rectifier with an \emph{angular RBF} kernel, and for regression, we use a fixed FIFO window of h=10 most recent features.

\noindent \textbf{Evaluation and Metrics.}
For text-to-image generation, we employ 200 DrawBench~\citep{saharia2022drawbench} prompts as inference inputs. To comprehensively assess the performance, we measure both generation fidelity and computational efficiency. We evaluate the visual quality of generated images using SSIM, PSNR, and LPIPS against the unaccelerated reference images, reporting the mean over all prompts. For efficiency evaluation, we report the actual inference latency in seconds and the theoretical computation cost in FLOPs, alongside their respective speedup ratios compared to the full-step baselines.

\subsection{Main Results}

\noindent \textbf{FLUX.1-dev} 
The quantitative results in Table~\ref{table:FLUX-Metrics} demonstrate that our method achieves a superior balance between acceleration and generation quality on the FLUX architecture. Fig. \ref{fig:compare} illustrates the visual comparison of various baseline models combined with our method running on FLUX. By integrating the Gaussian Process-based uncertainty quantification module into existing frameworks such as \textbf{TaylorSeer}, \textbf{HiCache}, and \textbf{ClusCa}, we significantly enhance image fidelity while maintaining approximately $3\times$ latency speedup and $5.8\times$ FLOPs reduction. Specifically, \textbf{ClusCa + ours} achieves the best PSNR of \textbf{29.823} and the lowest LPIPS of \textbf{0.3334} among all compared acceleration schemes. Compared to traditional step-reduction or baseline caching methods, the experimental data proves that our proposed refinement mechanism effectively mitigates error accumulation during the caching process.

\noindent \textbf{Qwen-Image}
We further evaluate the proposed method on the larger Qwen-Image model to verify its scalability. As shown in Table~\ref{table:Qwen-Metrics}, while the baseline (50 steps) requires a substantial latency of 36.68s, our GP-based refinement consistently achieves superior efficiency. Notably, the integration with \textbf{HiCache} and \textbf{TaylorSeer} enables a remarkable FLOPs speedup of over \textbf{6.2$\times$}, outperforming most caching-based baselines in both speed and fidelity. For instance, \textbf{TaylorSeer + ours} improves the PSNR from 28.20 to \textbf{29.48} ($+1.28$ dB) and significantly reduces the LPIPS from 0.65 to \textbf{0.29} compared to the $\mathcal{N}=7$ baseline. These results highlight that our method is particularly effective for large-scale models, where standard caching tends to accumulate errors more rapidly. By leveraging Gaussian Process priors, we recover critical visual features that are otherwise lost under extreme acceleration.

\subsection{Ablation Study}

\input{tables/Ablation-K}

\noindent \textbf{Ablation on GP Rectification and Dynamic Scheduling.}
We conduct a two-fold ablation study to verify the individual effectiveness of our proposed GP rectifier and the uncertainty-gated dynamic scheduling mechanism.

\textbf{First}, to evaluate the impact of \textbf{GP rectification}, we compare the baseline \textbf{TaylorSeer} with our integrated version (\textbf{TaylorSeer + ours}) under various fixed skip intervals $\mathcal{N} \in \{6, 7, 8\}$. As shown in Table~\ref{tab:dy_ablation}, the addition of our rectifier consistently yields substantial fidelity gains across all configurations; for instance, at $\mathcal{N}=7$, the PSNR improves from 28.671 to 29.266, while the LPIPS drops from 0.4542 to 0.4151. This demonstrates that the GP rectifier can effectively learn and compensate for the numerical residuals in the Taylor expansion features, regardless of the caching frequency. 

\textbf{Second}, we investigate the necessity of \textbf{dynamic scheduling}. We compare \textbf{TaylorSeer + Only Dynamic}, which utilizes Taylor features for interval adjustment without GP rectification, against our full model. While the Taylor-based dynamic approach achieves a $5.72\times$ speedup, its PSNR (28.930) is lower than even our fixed-interval version at $\mathcal{N}=6$. 

In contrast, our full configuration, \textbf{TaylorSeer + ours} ($O=2$), which leverages GP-derived uncertainty for both scheduling and rectification, achieves the best performance with a PSNR of \textbf{29.550} and a significant LPIPS reduction to \textbf{0.3497}. 

These results confirm that while both modules are independently effective, their synergy is crucial: the dynamic scheduling optimizes resource allocation based on model uncertainty, while the GP rectifier ensures that the approximated steps maintain high-frequency structural integrity.

% \section{Distribution of Dynamic Sampling Steps}
\noindent \textbf{Distribution of Dynamic Sampling Steps.}
\label{sec:appendix_nfe_distribution}
\begin{figure*}[h!]
    \centering
    \includegraphics[width=\textwidth]{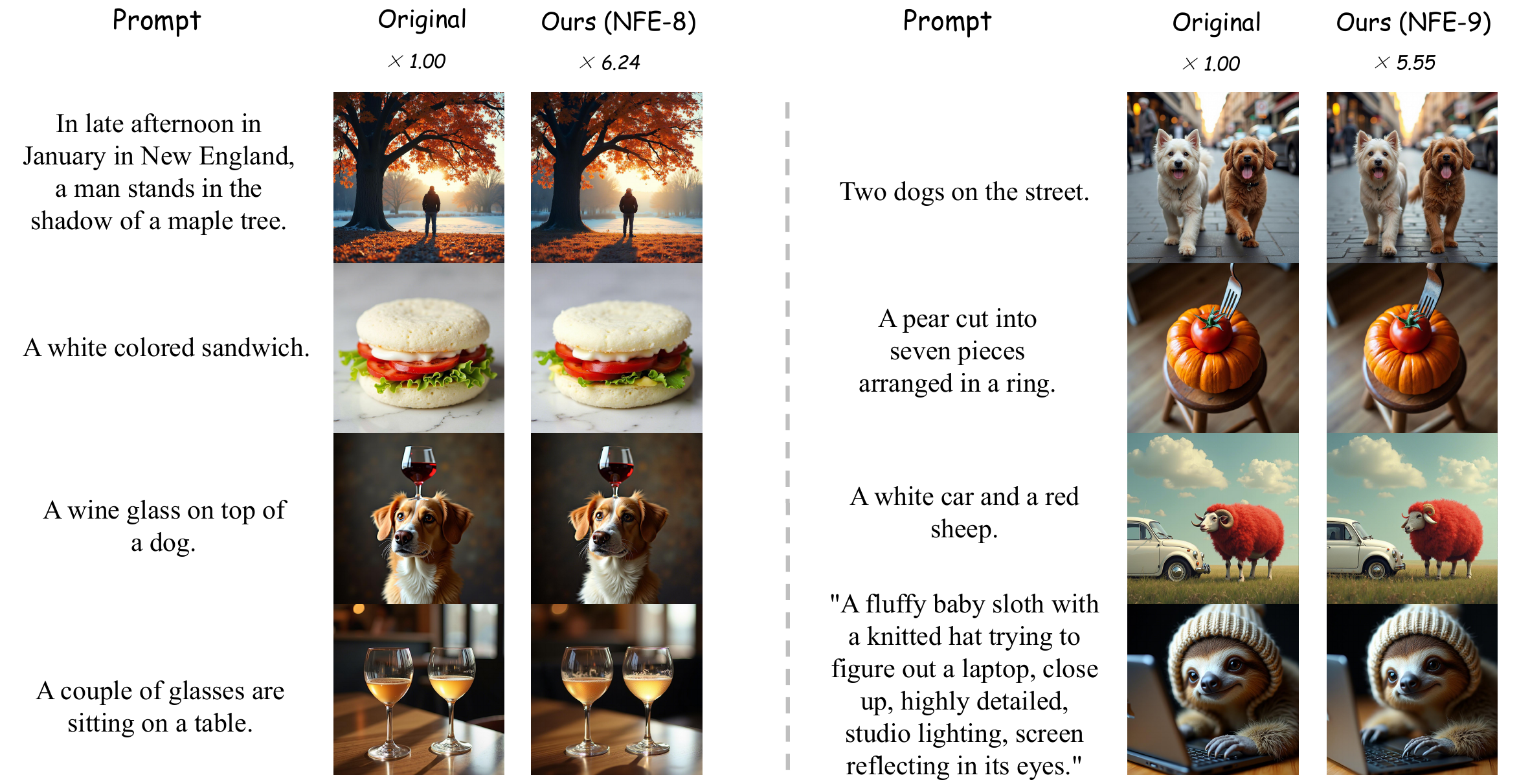} 
    \caption{\textbf{Qualitative results of GP-Refiner (TaylorSeer + ours) on FLUX at NFE-8 and NFE-9.}}
    \label{fig:qualitative_nfe}
    \vspace{-5mm}
\end{figure*}
To further investigate the stability and predictability of our dynamic interval scheduling strategy, we evaluated the distribution of the required Number of Function Evaluations (NFEs) on a diverse test set of \textbf{200 generated images} using the \textbf{TaylorSeer~\citep{liuTaylorSeer2025} + ours} configuration on the \textbf{FLUX~\citep{flux2024}} model. Experimental results demonstrate that \textbf{GP-Refiner} consistently converges within a \textbf{highly concentrated range} of steps. Specifically, among the 200 evaluated samples, \textbf{77} images required \textbf{8 NFEs}, \textbf{119} images required \textbf{8 NFEs}, and only \textbf{4} images required \textbf{9 NFEs}.

This highly concentrated NFE distribution proves that our stopping criterion possesses \textbf{exceptional stability}. It not only \textbf{avoid sdrastic fluctuations} in sampling steps during inference but also ensures that the overall computational overhead remains \textbf{strictly controllable} throughout the inference process.

Furthermore, as illustrated in Figure \ref{fig:qualitative_nfe}, the difference in visual quality between images generated with 9 NFEs and 10 NFEs is \textbf{marginal}. Our dynamic allocation mechanism successfully \textbf{prunes redundant computations} while perfectly preserving the \textbf{high fidelity} and aesthetic quality of the images.

\section{Conclusion}
Addressing the challenge of cumulative feature cache errors faced by diffusion transformers during accelerated inference, a plug-and-play feature rectification framework, GP-Refiner, is proposed. Driven by the statistical observation that the baseline prediction residuals exhibit a local zero-mean Gaussian distribution, the framework transforms the acceleration process into a joint task of baseline estimation and rectification. Furthermore, utilizing the posterior variance derived from the regression, an uncertainty-aware adaptive computation strategy is designed to dynamically trigger full-computation calibration when the inference confidence falls below a threshold. Extensive experiments on large-scale generative models demonstrate that this mechanism significantly enhances image fidelity while substantially reducing computational complexity and accelerating inference speed, thereby establishing a new benchmark for the trade-off between computational efficiency and generation quality.

\section{Acknowledgements}
This paper was partially sponsored by WUYING - Alibaba Cloud.

\bibliographystyle{splncs04}
\bibliography{main}
\end{document}

%% file: tables/FLUX-Metric.tex
\begin{table*}[ht]
    \centering
    \caption{\textbf{Quantitative comparison in text-to-image generation} for FLUX.}
    \vspace{3mm}
    \setlength\tabcolsep{8.0pt} 
      \normalsize
      \resizebox{0.99\textwidth}{!}{
      \begin{tabular}{l | c  c | c  c | c | c | c }
        \toprule
        {\bf Method} & {\bf Latency(s) $\downarrow$} & {\bf Speed $\uparrow$} & {\bf FLOPs(T) $\downarrow$}  & {\bf Speed $\uparrow$} & {\bf PSNR$\uparrow$} & {\bf SSIM$\uparrow$} & {\bf LPIPS$\downarrow$} \\
        \midrule
      
      $\textbf{[dev]: 50 steps}$  & {11.55}  & {1.00$\times$} & {3719.50}   & {1.00$\times$} & {-}  & {-}  & {-}      \\ 
      \midrule
    
      {$60\%$\textbf{ steps}}  & {7.12} & {1.62$\times$} & {2231.70} & {1.67$\times$} & {30.310} & {0.7819} & {0.2461}      \\
      {$50\%$\textbf{ steps}}  & {5.93} & {1.95$\times$} & {1859.75} & {2.00$\times$} & {29.576} & {0.7337} & {0.3106}             \\
      {$40\%$\textbf{ steps}}  & {4.82} & {2.40$\times$} & {1487.80} & {2.50$\times$} & {29.122} & {0.6971} & {0.3619}             \\
      {$34\%$\textbf{ steps}}  & {4.15} & {2.78$\times$} & {1264.63} & {2.94$\times$} & {28.881} & {0.6776} & {0.3913}             \\
      \midrule
      
      $\textbf{FORA}$ $(\mathcal{N}=5)$ & {3.31}  & {3.49$\times$} & {893.54} & {4.16$\times$} & {28.432}  & {0.6029}  & {0.4918}      \\
      $\textbf{FORA}$ $(\mathcal{N}=7)$ & {2.90}   & {3.98$\times$} & {670.44}   & {5.55$\times$} & {28.315} & {0.5870} &  {0.5409} \\
      $\textbf{\texttt{ToCa}}$ $(\mathcal{N}=6)$ & {7.28}   & {1.59$\times$} & {854.42}  & {4.35$\times$} & {29.135}  & {0.6532}  & {0.4081}             \\
      $\textbf{\texttt{ToCa}}$ $(\mathcal{N}=9)$ & {6.45}   & {1.79$\times$} & {784.54}   & {4.74$\times$} & {28.889} & {0.6407} &  {0.4525} \\
      $\textbf{\texttt{DuCa}}$ $(\mathcal{N}=8)$ & {3.46}   & {3.34$\times$} & {676.79}   & {5.50$\times$} & {29.133} & {0.6150} &  {0.4534} \\
      $\textbf{\texttt{DuCa}}$ $(\mathcal{N}=10)$ & {3.22}   & {3.59$\times$} & {606.91}   & {6.13$\times$} & {28.953} & {0.5957} &  {0.4935} \\
      $\textbf{TeaCache}$ $(l=0.6)$ & {3.76} & {3.07$\times$} & {1115.44} & {3.33$\times$} & {29.029}  & {0.6801}  & {0.4026}             \\
      $\textbf{TeaCache}$ $(l=1.0)$ & {2.77} & {4.17$\times$} & {743.63}   & {5.00$\times$} & {28.606} & {0.6360} &  {0.4773} \\
      
      \midrule
      
        $\textbf{TaylorSeer}$ $(O=2,\mathcal{N}=6)$ & {3.81}   & {3.03$\times$} & {744.80}   & {4.99$\times$} & {28.943} & {0.6558} & {0.4020} \\
        $\textbf{TaylorSeer}$ $(O=2,\mathcal{N}=7)$ & {3.60}   & {3.21$\times$} & {670.44}   & {5.55$\times$} & {28.671} & {0.6237} & {0.4542} \\
        \rowcolor{gray!20}
        $\textbf{TaylorSeer + ours}$ $(O=2)$ & {\textbf{3.42}}   & {\textbf{3.38}$\times$} & {\textbf{643.13}}  & {\textbf{5.78}$\times$} & {\textbf{29.550}} & {\textbf{0.6972}} & {\textbf{0.3497}} \\

      \midrule

        $\textbf{HiCache}$ $(\mathcal{N}=6)$ & {3.77}   & {3.06$\times$} & {744.80}  & {4.99$\times$} & {29.104} & {0.6685} & {0.3742} \\
        $\textbf{HiCache}$ $(\mathcal{N}=7)$ & {3.57}   & {3.23$\times$} & {670.44}  & {5.55$\times$} & {28.937} & {0.6572} & {0.3982} \\
        \rowcolor{gray!20}
        $\textbf{HiCache + ours}$ & {\textbf{3.45}} & {\textbf{3.35}$\times$} & {\textbf{631.33}} & {\textbf{5.89}$\times$} & {\textbf{29.694}} & {\textbf{0.6912}} & {\textbf{0.3881}} \\

      \midrule

        $\textbf{ClusCa}$ $(O=1,\mathcal{N}=6)$ & {4.09}   & {2.82$\times$} & {748.48} & {4.97$\times$} & {28.818} & {0.6557} & {0.4093} \\
        $\textbf{ClusCa}$ $(O=1,\mathcal{N}=7)$ & {3.88}   & {2.98$\times$} & {674.12} & {5.52$\times$} & {28.596} & {0.6276} & {0.4567} \\
        \rowcolor{gray!20}
        $\textbf{ClusCa + ours}$ $(O=1)$ & {\textbf{3.69}} & {\textbf{3.13}$\times$} & {\textbf{635.46}} & {\textbf{5.85}$\times$} & {\textbf{29.823}} & {\textbf{0.7100}} & {\textbf{0.3334}} \\
        \bottomrule
      \end{tabular}
      }
      
      \label{table:FLUX-Metrics}
    \end{table*}

%% file: tables/Qwen-Metric.tex
% 36.68
\begin{table*}[ht]
    \centering
    \caption{\textbf{Quantitative comparison in text-to-image gen.} for Qwen-Image.}
    \vspace{3mm}
    \setlength\tabcolsep{8.0pt} 
      \normalsize
      \resizebox{0.99\textwidth}{!}{
      \begin{tabular}{l | c  c | c  c | c | c | c}
        \toprule
        {\bf Method} & {\bf Latency(s) $\downarrow$} & {\bf Speed $\uparrow$} & {\bf FLOPs(T) $\downarrow$}  & {\bf Speed $\uparrow$} & {\bf PSNR$\uparrow$} & {\bf SSIM$\uparrow$} & {\bf LPIPS$\downarrow$} \\
        \midrule
      
      $\textbf{50 steps}$  & {36.68}  & {1.00$\times$} & {12917.56}   & {1.00$\times$} & {-}  & {-}  & {-}      \\ 
      \midrule
    
      % {$50\%$\textbf{ steps}}  & {} & {1.99$\times$} & {6458.78} & {2.00$\times$} & {30.54} & {0.75} & {0.28}      \\
      {$20\%$\textbf{ steps}}  & {7.82} & {4.69$\times$} & {2583.51} & {5.00$\times$} & {28.59} & {0.61} & {0.52}      \\
      
      \midrule

    \textbf{FORA} ($\mathcal{N}=4$) & 10.76 & 3.41$\times$ & 3359.99 & 3.84$\times$ & 28.66 & 0.59 & 0.51 \\
    \textbf{FORA} ($\mathcal{N}=6$) & 8.49 & 4.32$\times$ & 2326.74 & 5.55$\times$ & 28.48 & 0.55 & 0.59 \\

    \textbf{\texttt{ToCa}} ($\mathcal{N}=8$) & 16.82 & 2.18$\times$ & 2991.34 & 4.32$\times$ & 28.93 & 0.63 & 0.44 \\
    \textbf{\texttt{ToCa}} ($\mathcal{N}=12$) & 14.56 & 2.52$\times$ & 2406.20 & 5.37$\times$ & 28.69 & 0.57 & 0.53 \\

    \textbf{\texttt{DuCa}} ($\mathcal{N}=9$) & 10.27 & 3.57$\times$ & 2958.13 & 4.37$\times$ & 28.45 & 0.58 & 0.55 \\
    \textbf{\texttt{DuCa}} ($\mathcal{N}=12$) & 8.41 & 4.36$\times$ & 2171.56 & 5.95$\times$ & 28.38 & 0.57 & 0.60 \\
      
      \midrule
      
        $\textbf{TaylorSeer}$ $(\mathcal{N}=6)$ & {9.71}   & {3.78$\times$} & {2583.97}   & {5.00$\times$} & {28.58} & {0.62} & {0.46} \\
        $\textbf{TaylorSeer}$ $(\mathcal{N}=7)$ & {8.99}   & {4.08$\times$} & {2323.30}   & {5.56$\times$} & {28.20} & {0.48} & {0.65} \\
        \rowcolor{gray!20}
        $\textbf{TaylorSeer + ours}$ & {\textbf{8.37}}   & {\textbf{4.38}$\times$} & {\textbf{2085.26}}  & {\textbf{6.19}$\times$} & {\textbf{29.48}} & {\textbf{0.75}} & {\textbf{0.29}} \\

     \midrule
      
        $\textbf{HiCache}$ $(\mathcal{N}=6)$ & {9.88}   & {3.71$\times$} & {2583.97}   & {5.00$\times$} & {28.89} & {0.66} & {0.40} \\
        $\textbf{HiCache}$ $(\mathcal{N}=7)$ & {9.11}   & {4.03$\times$} & {2323.30}   & {5.56$\times$} & {28.64} & {0.63} & {0.45} \\
        \rowcolor{gray!20}
        $\textbf{HiCache + ours}$ & {\textbf{8.54}}   & {\textbf{4.30}$\times$} & {\textbf{2067.17}}  & {\textbf{6.25}$\times$} & {\textbf{29.39}} & {\textbf{0.74}} & {\textbf{0.32}} \\

    \midrule
    \textbf{Qwen-Image-Lightning-8steps}
    & 7.10 & 1.00$\times$ & 2123.17 & 1.00$\times$ & $\infty$ & 1.00 & 0.00 \\
    
    \textbf{TaylorSeer ($\mathcal{N}=3$)}
    & \textbf{4.56}
    & \textbf{1.56}$\times$
    & \textbf{1326.98}
    & \textbf{1.60}$\times$
    & 30.064
    & 0.6709
    & 0.2962 \\
    
    \rowcolor{gray!20}
    \textbf{TaylorSeer + ours ($\mathcal{N}=3$)}
    & 4.69
    & 1.51$\times$ 
    & 1326.99
    & 1.60$\times$
    & \textbf{31.204} 
    & \textbf{0.7665}
    & \textbf{0.1922} \\
      \bottomrule
      \end{tabular}
      }
      \label{table:Qwen-Metrics}
    \end{table*}

%% file: tables/Ablation-K.tex
\begin{table*}[ht]
    \centering
    \caption{\textbf{Ablation on GP Rectification and Dynamic Interval Scheduling.}}
    \vspace{3mm}
    \setlength\tabcolsep{8.0pt}
    \normalsize
    \resizebox{\textwidth}{!}{
    \begin{tabular}{l | c | c | c | c | c}
        \toprule
        {\bf Method} & {\bf FLOPs(T) $\downarrow$} & {\bf Speed $\uparrow$} & {\bf PSNR$\uparrow$} & {\bf SSIM$\uparrow$} & {\bf LPIPS$\downarrow$} \\
        \midrule

        $\textbf{[dev]: 50 steps}$ & {3719.50} & {1.00$\times$} & {-} & {-} & {-} \\
        \midrule

        $\textbf{TaylorSeer}$ $(O=2, \mathcal{N}=8)$ & {596.07} & {$6.24\times$} & {28.513\textcolor{gray!70}{\scriptsize (+0.00)}} & {0.5986\textcolor{gray!70}{\scriptsize (+0.00)}} & {0.4854\textcolor{gray!70}{\scriptsize (+0.00)}} \\
        $\textbf{TaylorSeer + ours}$ $(O=2, \mathcal{N}=8)$ & {596.08} & {$6.24\times$} & {29.129\textcolor{gray!70}{\scriptsize (+0.62)}} & {0.6426\textcolor{gray!70}{\scriptsize (+0.04)}} & {0.4469\textcolor{gray!70}{\scriptsize (-0.04)}} \\
        \midrule

        $\textbf{TaylorSeer}$ $(O=2, \mathcal{N}=7)$ & {670.44} & {$5.55\times$} & {28.671\textcolor{gray!70}{\scriptsize (+0.00)}} & {0.6237\textcolor{gray!70}{\scriptsize (+0.00)}} & {0.4542\textcolor{gray!70}{\scriptsize (+0.00)}} \\
        $\textbf{TaylorSeer + ours}$ $(O=2, \mathcal{N}=7)$ & {670.45} & {$5.55\times$} & {29.266\textcolor{gray!70}{\scriptsize (+0.60)}} & {0.6622\textcolor{gray!70}{\scriptsize (+0.04)}} & {0.4151\textcolor{gray!70}{\scriptsize (-0.04)}} \\
        \midrule

        $\textbf{TaylorSeer}$ $(O=2, \mathcal{N}=6)$ & {744.80} & {$4.99\times$} & {28.943\textcolor{gray!70}{\scriptsize (+0.00)}} & {0.6558\textcolor{gray!70}{\scriptsize (+0.00)}} & {0.4020\textcolor{gray!70}{\scriptsize (+0.00)}} \\
        $\textbf{TaylorSeer + ours}$ $(O=2, \mathcal{N}=6)$ & {744.81} & {$4.99\times$} & {29.432\textcolor{gray!70}{\scriptsize (+0.49)}} & {0.6746\textcolor{gray!70}{\scriptsize (+0.02)}} & {0.3895\textcolor{gray!70}{\scriptsize (-0.01)}} \\
        \midrule

        $\textbf{TaylorSeer + Only Dynamic}$ $(O=2)$ & {650.21} & {$5.72\times$} & {28.930\textcolor{gray!70}{\scriptsize (+0.00)}} & {0.6604\textcolor{gray!70}{\scriptsize (+0.00)}} & {0.3819\textcolor{gray!70}{\scriptsize (+0.00)}} \\
        \rowcolor{gray!20}
        $\textbf{TaylorSeer + ours}$ $(O=2)$ & {\textbf{643.13}}  & {$\textbf{5.78}\times$} & {\textbf{29.550}\textcolor{gray!70}{\scriptsize (+0.62)}} & {\textbf{0.6972}\textcolor{gray!70}{\scriptsize (+0.04)}} & {\textbf{0.3497}\textcolor{gray!70}{\scriptsize (-0.03)}} \\

        \bottomrule
    \end{tabular}
    }
    \label{tab:dy_ablation}
\end{table*}